\documentclass[final]{cvpr}

\usepackage{times}
\usepackage{epsfig}
\usepackage{graphicx}
\usepackage{amsmath}
\usepackage{amssymb}
\usepackage{microtype}
\usepackage{bm}
\usepackage{subfigure}

\usepackage{enumitem}
\setitemize[0]{leftmargin=15pt}

\newenvironment{tight_itemize}{
\begin{itemize}[leftmargin=10pt]
 \setlength{\topsep}{0pt}
 \setlength{\itemsep}{2pt}
 \setlength{\parskip}{0pt}
 \setlength{\parsep}{0pt}
}{\end{itemize}}

\usepackage[pagebackref=true,breaklinks=true,colorlinks,bookmarks=false]{hyperref}

\def\cvprPaperID{1678} 
\def\confYear{CVPR 2021}

\begin{document}
\title{Fusing the Old with the New: \\Learning Relative Camera Pose with Geometry-Guided Uncertainty}

\author{Bingbing Zhuang$^1$\\
$^1$NEC Labs America\\
\and
Manmohan Chandraker$^{1,2}$\\
$^2$University of California, San Diego\\
}

\maketitle

\begin{abstract}
Learning methods for relative camera pose estimation have been developed largely in isolation from classical geometric approaches. The question of how to integrate predictions from deep neural networks (DNNs) and solutions from geometric solvers, such as the 5-point algorithm~\cite{nister2004efficient}, has as yet remained under-explored. In this paper, we present a novel framework that involves probabilistic fusion between the two families of predictions during network training, with a view to leveraging their complementary benefits in a learnable way. The fusion is achieved by learning the DNN uncertainty under explicit guidance by the geometric uncertainty, thereby learning to take into account the geometric solution in relation to the DNN prediction. Our network features a self-attention graph neural network, which drives the learning by enforcing strong interactions between different correspondences and potentially modeling complex relationships between points. We propose motion parmeterizations suitable for learning and show that our method achieves state-of-the-art performance on the challenging DeMoN~\cite{ummenhofer2017demon} and ScanNet~\cite{dai2017scannet} datasets. While we focus on relative pose, we envision that our pipeline is broadly applicable for fusing classical geometry and deep learning.
\end{abstract}

\section{Introduction}
\label{sec:intro}

Estimating the relative pose between two cameras is a fundamental problem in computer vision, which forms the backbone of structure from motion (SFM) methods. Geometric approaches based on the 5-point method~\cite{nister2004efficient} and bundle adjustment (BA)~\cite{triggs1999bundle} are well-studied, while recent methods based on deep neural networks (DNNs) also achieve promising results~\cite{ummenhofer2017demon,clark2018ls,tang2018ba,wei2019deepsfm}. But the question of how the two families of methods may be combined to trade-off their relative benefits has as yet remained under-explored, which is the subject of our study in this paper.

The behavior of geometric methods~\cite{hartley2003multiple} is theoretically characterizable under a wide range of camera motions. But such understanding does not always guarantee good performance. While high accuracy is obtained in situations with strong perspective effects, performance may degrade due to lack of correspondences, planar degeneracy and bas-relief ambiguity~\cite{daniilidis1997understanding}, to name a few. On the other hand, learning-based methods may avoid the above issues by learning sophisticated priors that relate images to camera motion, but can suffer from poor generalization outside the training domain and not be amenable to interpretation. 

\begin{figure}[t]
  \centering
  \includegraphics[width=1.0\linewidth, trim = 0mm 110mm 210mm 0mm, clip]{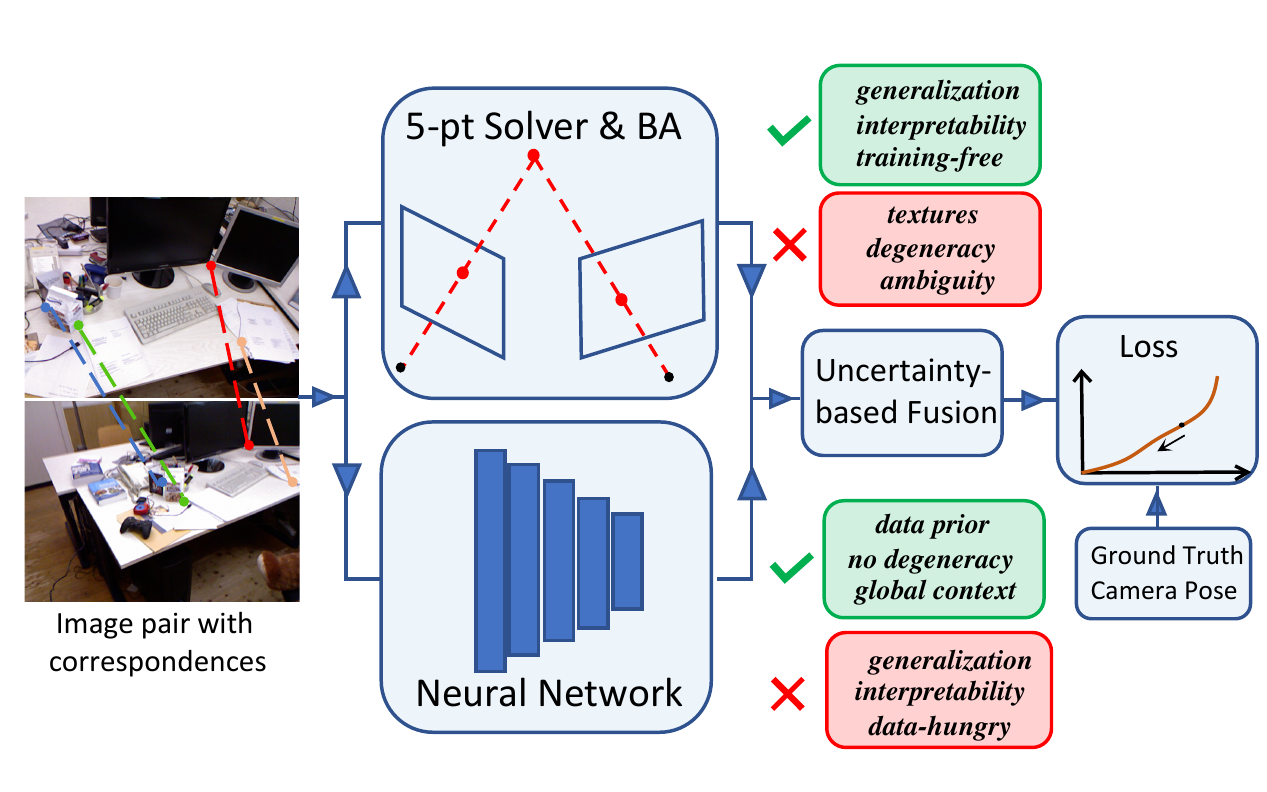}
  \centering
  \caption{\textbf{Geometric-DNN relative pose fusion framework.} The DNN pose prediction is fused with the geometric prediction during training, based on their respective prediction uncertainty.}
  \label{fig:teaser}
\end{figure}

This paper proposes an uncertainty based probabilistic framework to fuse geometric and DNN predictions, as illustrated in Fig.~\ref{fig:teaser}, with the aim of overcoming the limitations of either approach. The underlying intuition is that the geometric solution may be trusted more due to its well-understood rationale if it is highly confident, but the network should play a role in driving the solution closer to the true one in geometrically ill-conditioned scenarios. We obtain geometric uncertainty using the Jacobian of the error functions, serving as an indicator of the quality of the solution, while we design a network to additionally predict the uncertainty associated with camera pose estimation. The uncertainty so obtained may be interpreted as (co)variance of a Gaussian distribution, which allows us to fuse the two predictions using Bayes' rule. We highlight that the geometric solution and the fusion step are both tightly integrated into our end-to-end trainable pipeline, hence enforcing strong interaction between DNNs and the geometric method during training.  

Our relative pose learning framework is the first of its kind in terms of forcing the network to give an account of the classical geometric solution along with its uncertainty in a principled way, during training. The network can also be thought of as a means to learn to improve the geometric solution such that the final fused one is closer to the ground truth. Further, the geometric guidance also distinguishes our uncertainty learning from previous works (e.g.~\cite{klodt2018supervising}) that learn standard aleatoric uncertainty~\cite{kendall2017uncertainties}. More importantly, our learned uncertainty may be considered as geometrically calibrated, in the sense that its numerical range can readily match to that of the geometric uncertainty and permits a direct fusion of the two during training. 

In terms of network architecture, inspired by SuperGlue \cite{sarlin2020superglue}, we find a self-attention~\cite{vaswani2017attention} graph neural network (GNN) to be effective at learning from keypoint correspondences. This is probably since self-attention permits strong interactions between correspondences, which is an essential procedure to determine the relative pose.  We also illustrate that even both translation direction and rotation lie on a manifold, fusion is still feasible by careful choice of parameterization. We term our uncertainty-aware fusion framework as UA-Fusion. UA-Fusion is extensively validated by achieving state-of-the-art performance, especially in challenging indoor datasets with unconstrained motions.   


In summary, our contributions include:
\vspace{-0.2cm}
\begin{itemize}
\item A principled fusion framework to leverage the best of both classical geometric solvers and DNNs for relative pose estimation. 
\vspace{-0.2cm}
\item A self-attention graph neural network whose attention mechanism drives the learning in our fusion pipeline.
\vspace{-0.2cm}
\item Superior results on benchmark DeMoN dataset~\cite{ummenhofer2017demon} as well as in cross-dataset ScanNet experiments \cite{dai2017scannet}. 
\end{itemize}

\section{Related Works}
\noindent \textbf{Geometric methods~~~} 
Due to its essential role in SFM~\cite{schonberger2016structure,zhu2018very,zhuang2018baseline}, a wide variety of algorithms have been proposed for the relative pose estimation problem. These could be classified into algebra-based minimal/nonminimal solvers~\cite{longuet1981computer,nister2004efficient,hartley1997defense,li2006five,kukelova2013algebraic} and optimization-based nonlinear methods~\cite{kneip2013direct,yang2014optimal,Briales_2018_CVPR,Fredriksson_2016_CVPR}. In addition, some methods are specially tailored to specific types of camera setup or motion prior~\cite{zhao2020certifiably,hee2013motion,zhuang2017rolling,hajder2020relative,sweeney2014solving,guan2020minimal,guan2020relative}. 
Our framework does not require the geometric methods to be differentiable, and in principle could work with any approaches.

\vspace{0.2cm}
\noindent \textbf{Learning methods~~~} Deep learning has recently been applied to different problems in SFM, e.g.~\cite{ranftl2018deep,sattler2019understanding,tiwari2020pseudo,tan2020self,zhuang2019degeneracy}.
In particular, both supervised~\cite{ummenhofer2017demon,clark2018ls,tang2018ba,wei2019deepsfm} and unsupervised~\cite{zhou2017unsupervised,wang2018learning,mahjourian2018unsupervised,bian2019depth,zhou2019moving} methods for relative pose estimation are proposed. Despite the promising performance, such methods are largely developed in isolation of geometric methods. Although many works do borrow ideas from geometric methods to guide the network designing, (e.g. the spirit of bundle adjustment in BA-Net~\cite{tang2018ba}, LS-Net~\cite{clark2018ls} and DeepSfM~\cite{wei2019deepsfm}), the geometric solution \textit{itself} is not explicitly leveraged in network training as we do. Note that while we validate our concepts by learning the camera pose alone, our idea is amenable to learning with depth as well.

\noindent \textbf{Geometric uncertainty~~~}
In addition to the well-behaved camera pose estimation, geometric methods also offer a natural way to measure the uncertainty~\cite{forstner2016photogrammetric} of the predictions without requiring access to the ground truth. This is achieved by relating the Jacobian of the error landscape to the covariance of Gaussian distributions. The uncertainty so obtained has been extensively studied in photometric computer vision~\cite{forstner2016photogrammetric}. It has also found applicability in a wide variety of tasks, such as RGB-D SLAM~\cite{fontan2020information}, radial distortion model selection~\cite{polic2020uncertainty} in SFM, skeletal images selection for efficient SFM~\cite{snavely2008skeletal}, height map fusion~\cite{zienkiewicz2016real}, 3D reconstruction~\cite{haner2012covariance}, and camera calibration~\cite{peng2019calibration}. Polic et al. recently make efforts~\cite{polic2017camera,polic2018fast} towards efficient uncertainty computation in large-scale 3D reconstruction.
Despite the prevalence of the geometric uncertainty, it is however not yet fully exploited when it comes to the context of deep learning.
Our work makes contributions towards bridging this gap.

\noindent \textbf{Learning uncertainty~~~}
Uncertainty learning, as a means towards more interpretable deep learning models, has emerged as an important topic recently~\cite{kendall2017uncertainties}. Quantifying uncertainty of network predictions is highly desirable in many applications, such as camera relocalization~\cite{kendall2016modelling}, depth uncertainty~\cite{bloesch2018codeslam} and photometric uncertainty~\cite{yang2020d3vo} in SLAM, and optical flow estimation~\cite{ilg2018uncertainty}. In the context of SFM, Klodt and Vedaldi~\cite{klodt2018supervising} utilize uncertainty learning to handle the varying reliability of geometric SFM when applied as reference ground truth to supervise the SFM learning. Our paper is close to but distinct from this work since our goal lies in the fusion between the geometric and learned SFM with both the geometric and learned uncertainty. Laidlow et al.~\cite{laidlow2019deepfusion,laidlow2020towards} put forth to fuse the depth maps obtained from learning and geometric methods, sharing similar spirit to our pose fusion framework. Yet, their uncertainty is learned independent from the geometric solution, and requires a non-trivial post-processing fusion step. In contrast, our UA-Fusion learns uncertainty by tightly integrating geometric guidance and fusion into network training.

\section{Method}
\label{sec:method}


\vspace{-0.0cm}
\paragraph{Tradeoffs between geometric and learned pose}
We start by noting intuition from geometric pose estimation, which may guide regimes where interesting trade-offs may be observed for our uncertainty-based fusion.


\vspace{-0.2cm}
\begin{tight_itemize}
\item {\em Critical keypoint configurations:} Geometric solvers suffer when correspondences are scarce, for example, in textureless regions. Some common keypoint configurations may lead to ambiguous solutions or ill-posed objectives, for example, when keypoints lie on a plane~\cite{hartley2003multiple}. We expect that our geometric uncertainty will lend greater importance to DNN priors in such situations.

\item {\em Critical motions:} Bas-relief ambiguity \cite{daniilidis1997understanding,szeliski1997shape} may arise when the camera undergoes sideways motion due to the resemblance between translational and rotational flow under limited field of view, leading to a less accurate pose estimation. Forward motion also poses challenges to geometric SFM, partially due to small feature movement near the focus of expansion and partially due to severe local minima in the least squares error landscape \cite{vedaldi2007moving,oliensis2005least}.

\item {\em Rotation and translation:} Translation estimates are known to be more sensitive than rotation, leading to issues such as forward motion bias in linear methods if proper normalization is not carried out \cite{nister2004efficient,tian1996comparison}. Thus, one may expect rotation to be more reliable in the geometric solution, while the DNN may play a more significant role for translation. We will show that our uncertainty-based framework handles this in a principled way.
\end{tight_itemize}

\subsection{Background: Geometric Uncertainty}

\paragraph{Geometric solution} Formally, we are interested in solving the relative camera pose between two cameras $C_1$ and $C_2$ with known intrinsics. We assume $C_1$ as the reference with pose denoted as $\boldsymbol{P}_1=[\boldsymbol{I}~~\boldsymbol{0}]$ and wish to estimate the relative camera pose of $C_2$, denoted as $\boldsymbol{P}_2=[\boldsymbol{R}~~\boldsymbol{t}]$, where $\boldsymbol{R} \in SO(3)$ and $t\in {\mathcal{S}}^2$ denote the relative rotation and translation direction, respectively. Suppose both cameras are viewing a set of common 3D points $\boldsymbol{X}_i$, $i=1,2,...,n$, each yielding a pair of 2D correspondences $\boldsymbol{x}_i^1$ and $\boldsymbol{x}_i^2$ in the image plane. It is well-known that a minimal set of 5-point correspondences suffices to determine the solution, with Nister's 5-point algorithm~\cite{nister2004efficient} being the standard minimal solver. A RANSAC procedure is usually applied to obtain an initial solution, followed by triangulation~\cite{hartley2003multiple} to obtain 3D points $\boldsymbol{X}_i$. Finally, one could refine the solution by nonlinearly minimizing the re-projection error using bundle adjustment~\cite{triggs1999bundle},
\begin{equation}
    \min_{\boldsymbol{\theta}}\sum_{i}\|\boldsymbol{x}_i^1-\pi(\boldsymbol{P}_1,\boldsymbol{X}_i))\|^2+\|\boldsymbol{x}_i^2-\pi(\boldsymbol{P}_2,\boldsymbol{X}_i))\|^2,
    \label{eq:ba}
\end{equation}
where $\pi()$ denotes the standard perspective projection and  $\boldsymbol{\theta}=\{\boldsymbol{\theta}_R,~\boldsymbol{\theta}_t,\boldsymbol{X}_i, i=1,2,...n\}$. $\boldsymbol{\theta}_R$ and $\boldsymbol{\theta}_t$ represent the parameterizations of rotation and translation; we will come back to their specific choice in the next section.

\vspace{-0.3cm}
\paragraph{Geometric uncertainty} In order to describe the uncertainty associated with the optimum $\hat{\boldsymbol{\theta}}$ in a probabilistic manner, the distribution of $\boldsymbol{\theta}$ could be approximated locally by a Gaussian distribution $\mathcal{N}(\boldsymbol{\theta}|\hat{\boldsymbol{\theta}},\boldsymbol{\Sigma})$. As a first-order approximation, the information matrix $\boldsymbol{\Lambda}$, i.e. $\boldsymbol{\Sigma}^{-1}$, is computed as:
\begin{equation}
    \boldsymbol{\Lambda}=J^\top(\hat{\boldsymbol{\theta}})J(\hat{\boldsymbol{\theta}}),
\end{equation}
where $J(\hat{\boldsymbol{\theta}})$ denotes the Jacobian of the nonlinear least squares (Eq.~\ref{eq:ba}) at $\hat{\boldsymbol{\theta}}$. We remark that $J(\hat{\boldsymbol{\theta}})$ is of full rank in this paper, implying the absence of gauge ambiguity~\cite{forstner2016photogrammetric,kanatani2001gauges}. This is attributed to the fixed camera pose of $C_1$ as well as our minimal parameterizations of $(\boldsymbol{R},\boldsymbol{t})$ to be discussed shortly. Also, we shall conduct fusion on each individual parameter in $\{\boldsymbol{\theta}_R,~\boldsymbol{\theta}_t\}$ separately due to the discontinuity~\cite{zhou2019continuity} in representation, and we will directly work on inverse variance for convenience. The inverse variance $1/\sigma_i^{2}$ of a parameter $\theta_i$ in $\{\boldsymbol{\theta}_R,~\boldsymbol{\theta}_t\}$ may be obtained by Schur complement:
\begin{equation}
    1/\sigma_i^{2} = \boldsymbol{\Lambda}\setminus\boldsymbol{\Lambda}_{J,J} = \boldsymbol{\Lambda}_{i,i} - \boldsymbol{\Lambda}_{i,J}\boldsymbol{\Lambda}_{J,J}^{-1}\boldsymbol{\Lambda}_{J,i},
\end{equation}
where $J$ includes the index to the remaining parameters in $\boldsymbol{\theta}$. This step is also called S-transformation~\cite{baarda1973s} that specifies the gauge of covariance matrix~\cite{forstner2016photogrammetric,polic2020uncertainty}. From the probabilistic point of view, it is in essence the inverse variance of a conditional Gaussian on $\theta_i$ given all the other parameters~\cite{johnson2002applied}. 
As the expert reader may have noticed, we omit the keypoint localization uncertainty~\cite{forstner2016photogrammetric,peng2019calibration} of $\boldsymbol{x}_i^{1,2}$, for simplicity.

\subsection{Geometric-DNN Relative Pose Fusion}
\noindent \textbf{Notations~~~}Whenever ambiguity arises, we use subscript g, d, and f to distinguish the prediction from the geometric prediction (g), the DNN prediction prior to fusion (d), and the fused solution (f).

\begin{figure*}[t]
	\centering
	    \vspace{-0.2cm}
		\includegraphics[width=1.0\linewidth, trim = 0mm 15mm 0mm 0mm, clip]{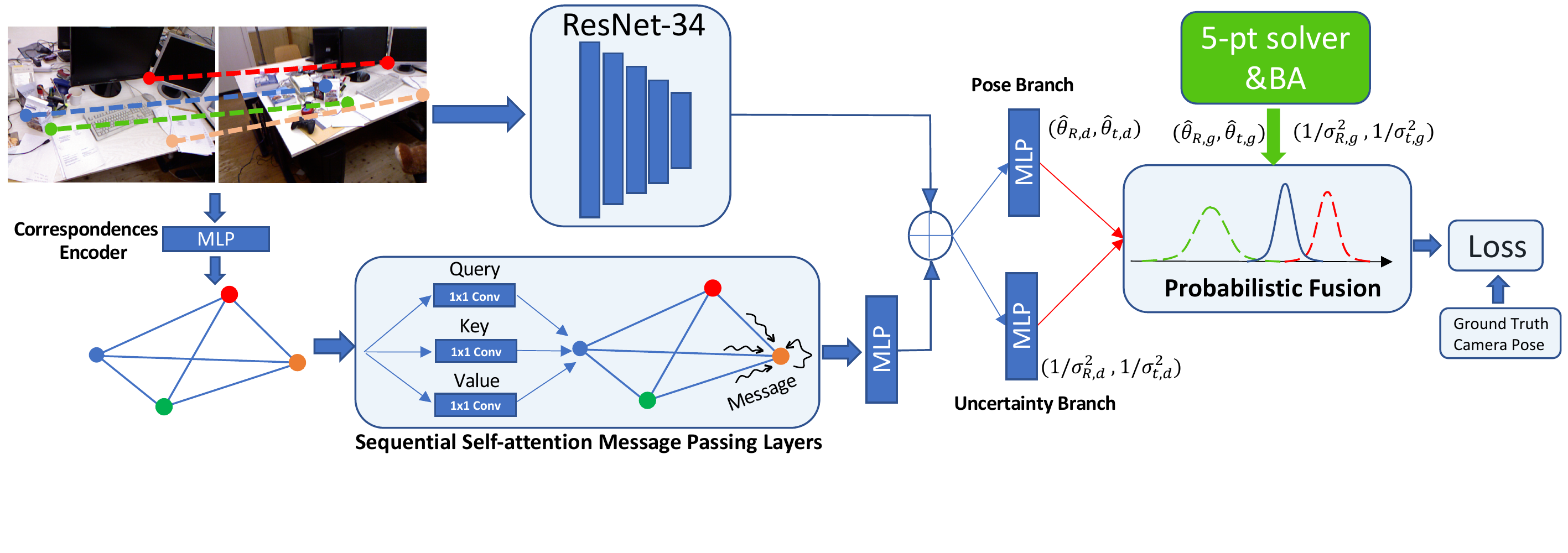}
	\centering
	\caption{\textbf{Overview of our geometric-DNN fusion network}, taking two images with extracted keypoint correspondences as input. The images and correspondences are respectively passed to the ResNet-34 and the self-attention graph network for appearance and geometric feature extraction. The concatenated features are passed to the two-branch MLPs for estimating pose and uncertainty. This solution is then fused with the one from the 5pt\&BA method with Jacobian-based uncertainty, yielding the final prediction which receives supervision.}
	\label{fig:pipeline}
\end{figure*}

\noindent \textbf{Bayes fusion~~~}We conceptually treat the geometric and DNN predictions as measurements from two different sensors and fuse them using Bayes' rule, akin to a Kalman filter. Specifically, given $(\hat\theta_{i,g},1/\sigma_{i,g}^2)$ and $(\hat\theta_{i,d},1/\sigma_{i,d}^2)$ as the geometric and DNN prediction, respectively, the posterior distribution of the motion parameter ${\theta}_i$
is~\cite{murphy2012machine} 
\begin{equation}
  P({\theta}_i|\hat{{\theta}}_{i,g},{\sigma}_{i,g}^2,\hat{{\theta}}_{i,d},{\sigma}_{i,d}^2) = \mathcal{N}({\theta}_i|\hat{{\theta}}_{i,f},\sigma_{i,f}^2),
\end{equation}
\begin{equation}
\hat{{\theta}}_{i,f} = \frac{1/\sigma_{i,g}^{2}~\hat{{\theta}}_{i,g} +1/\sigma_{i,d}^{2}~\hat{{\theta}}_{i,d}}{1/\sigma_{i,g}^{2}+1/\sigma_{i,d}^{2}},~1/\sigma_{i,f}^2 =(1/\sigma_{i,g}^{2}+1/\sigma_{i,d}^{2}).    
\end{equation}
The maximum-a-posterior (MAP) estimation $\hat{{\theta}}_{i,f}$ is returned as the final fused solution, which receives supervision.

\vspace{-0.2cm}
\subsubsection{Neural Architecture for Probabilistic Fusion}
\label{sec:architecture}

\noindent{\textbf{Architecture Overview}~~~}
As illustrated in Fig.~\ref{fig:pipeline}, our UA-Fusion framework takes as input an image pair along with the correspondences extracted by a feature matcher, for which we use SuperGlue~\cite{sarlin2020superglue} due to its excellent performance. The two images are stacked and passed to a ResNet~\cite{he2016deep} architecture to extract the appearance feature. The corresponding keypoint locations are first embeded into a higher-dimensional space by a Multilayer Perceptron (MLP) and then feed into an attentional graph neural network to extract the geometric feature. Afterwards, the appearance and geometric features are concatenated before being passed to the pose and the uncertainty branches, each using an MLP to predict respectively the mean and inverse variance of the underlying Gaussian distribution of the motion parameters. These are then fused with the geometric solution (5pt\&BA) based on uncertainty. Note that the loss is imposed on the final fused output, which induces gradient flows back-propagated through the fusion, hence coupling the geometric and learning module in training. 

\noindent{\textbf{Intuition for the Architecture}~~~}While the ResNet offers global appearance context, the graph neural network and geometric feature encode strong geometric cues from any available correspondences to reason about camera motion. Further, correspondences as the sole input to the 5-point solver and bundle adjustment have a more explicit correlation with the uncertainty of geometric solution, allowing the network to decide the extent to which the geometric solution should be trusted in relation to the DNN prediction.

\noindent{\textbf{Self-Attention Graph Neural Network}~~~}
As the network input, we stack all the correspondences $(\boldsymbol{x}_i^1,\boldsymbol{x}_i^2)$ between the two views as $\boldsymbol{x}^{12} \in R^{n\times4}$, which is subsequently passed to an MLP for embedding, yielding $\boldsymbol{f}^{(0)} \in R^{n\times d}$, with $d=128$ being the feature dimension. Next, $\boldsymbol{f}^{(0)}$ is passed to four sequential message passing layers to propagate information between all pairs of correspondences, with structure similar to SuperGlue~\cite{sarlin2020superglue}. The message passing is best represented as a graph, with each node containing a pair of correspondence and edges connecting different pairs of correspondences.  Specifically, in the $l$-th layer, the feature vector $\boldsymbol{f}_i^{l}$ associated with the correspondence pair $i$ is updated as:
\begin{equation}
    \boldsymbol{f}_i^{l+1} = \boldsymbol{f}_i^{l} + \text{MLP}([\boldsymbol{f}_i^{l},\boldsymbol{m}_i^l]),
\end{equation}
where $[~.~,~.~]$ indicates concatenation and $\boldsymbol{m}_i^l$ denotes the message aggregated from all the correspondences based on the self-attention mechanism~\cite{vaswani2017attention}. As per the standard procedure, we define the query ($\boldsymbol{Q}^l$), key ($\boldsymbol{K}^l$) and value ($\boldsymbol{V}^l$) as linear projections of $\boldsymbol{f}^l$, each with their own learnable parameters shared across all the correspondences. The message $\boldsymbol{m}^l$ is then computed as
\begin{equation}
    \boldsymbol{m}^l = \text{softmax}(\frac{\boldsymbol{Q}^l{\boldsymbol{K}^l}^\top}{\sqrt{d}})\boldsymbol{V}^l,
\end{equation}
The softmax is performed row-wise and $\boldsymbol{m}_i^l$ is row $i$ of $\boldsymbol{m}^l$. The output of the last layer $\boldsymbol{f}_i^4$ is passed to an MLP and average pooling to be concatenated with the ResNet feature.

\noindent{\textbf{Why self-attention?}~~~} First of all, the self-attention encourages interactions between all correspondences, which mimics the knowledge from classical geometry that all pairs of correspondences ($n>=5$) together contribute to the relative pose determination, necessitating the interactions between points. More importantly, the strong representation capability of self-attention facilitates the learning of complex relationships among different pairs of correspondences. A straightforward example is the spatial relation. It is known in classical SFM \cite{hartley2003multiple} that two pairs of correspondences far from each other and widely spread in the image plane provide stronger signals for motion estimation; it effectively makes full use of the perspective effect in the field of view and prevents the degradation of perspective to affine camera model~\cite{hartley2003multiple,shapiro19953d}. Conversely, two pairs of correspondences near to each other typically contribute weaker extra cues compared to either pair alone. 
Hence, different pairs of correspondences are not on equal footing and should be treated differently.  Indeed, we empirically observe a strong correlation between the attention and the spatial distance.
In a nutshell, the self-attention enforces extensive interactions and permits learning  more complex and abstract relationships among different pairs of correspondences, which we empirically observe to play a significant role in terms of performance. More analyses will follow in the experiments.

\subsubsection{Motion parameterization}
\label{sec:motionparameter}
It is crucial to choose the proper motion parameterizaton for the above network, which we discuss now.

\noindent{\textbf{Translation~~~}}We consider the properties of two distinct parameterizations for training and fusion:
\begin{equation}
    \boldsymbol{t}(t_x,t_y,t_z) = \frac{(t_x,t_y,t_z)^\top}{\|(t_x,t_y,t_z)^\top\|},
\end{equation}
\begin{equation}
 \boldsymbol{t}(\alpha,\beta) = (\cos\alpha, \sin\alpha\cos\beta , \sin\alpha\sin\beta),
\end{equation}
where $\alpha \in [0,\pi]$ and $\beta \in [-\pi,\pi]$ are constraints for uniqueness. Since the unit-norm vector $\boldsymbol{t}$ itself is not a convenient quantity for fusion as it lies on the ${\mathcal{S}}^2$ manifold, we seek to fuse the parameters $(t_x,t_y,t_z)$ or $(\alpha,\beta)$. As the scale of $(t_x,t_y,t_z)$ is indeterminate, causing gauge ambiguity and a rank-deficient Jacobian, we opt for $(\alpha,\beta)$ as the fusion entity, that is, $\boldsymbol{\theta}_t=\{\alpha,\beta\}$. However, due to its circular nature, the wrap-around of $\beta$ at $\pm\pi$ leads to discontinuity in the representation, which leads to training difficulties if the network predicts $\beta$ directly. To address this issue, we design the network to output $(t_x,t_y,t_z)$ followed by unit-normalization, then we extract $(\alpha,\beta)$ and proceed to fusion. 

\begin{figure}[t]
  \centering
  \includegraphics[width=0.7\linewidth, trim = 0mm 16mm 0mm 0mm, clip]{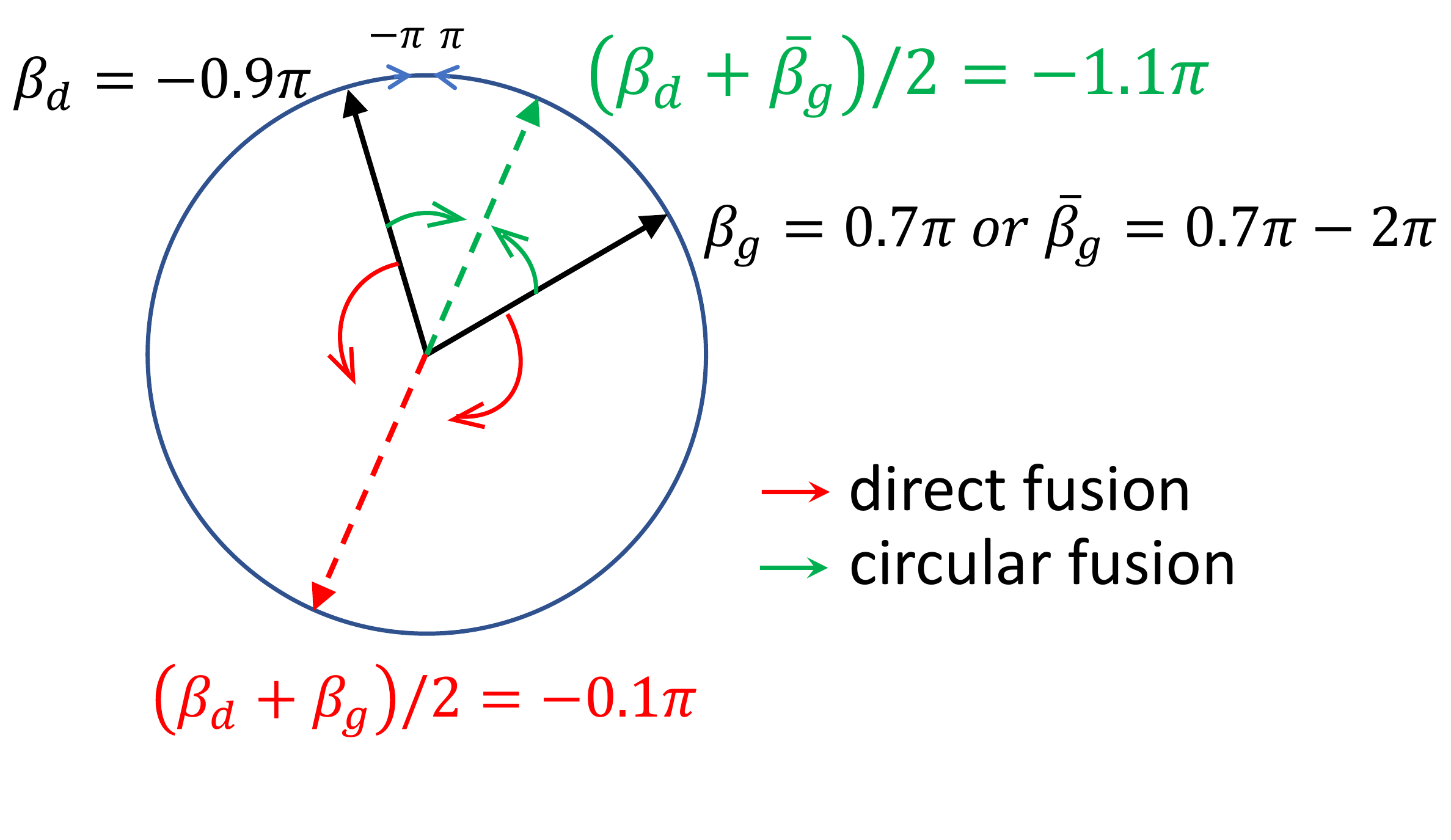}
  \centering
  \caption{\textbf{Toy-example illustration} of the direct and circular fusion on $\beta$, wherein the fusion is simply defined as averaging.}
  \label{fig:fusiontoyexample}
\end{figure}

\noindent{\textbf{Circular Fusion}~~~}While the fusion of $\alpha$ remains straightforward, the circular nature slightly complicates the fusion of $\beta$. Ideally, a meaningful fusion is obtained only when $|\beta_d-\beta_g|<\pi$, which could be achieved by letting $\bar\beta_g=\beta_g+2k\pi$ with $k \in \{-1,0,1\}$. This is illustrated by the toy example in Fig.~\ref{fig:fusiontoyexample}, where depending upon the specific values of $\beta_g$ and $\beta_d$, a direct fusion of the two might yield a solution far from both when $|\beta_d-\beta_g|>\pi$. This is, however, addressed by fusing $\beta_d$ and $\bar\beta_g$ instead. We term this procedure as circular fusion and $\bar\beta_g$ as $\beta_g$'s circular nearest neighbor to $\beta_d$. 

\noindent{\textbf{Rotation}~~~}We consider two minimal 3-parameter representation of rotation---angle-axis representation and Euler angles. The network is designed to regress the angle directly. Although this also faces discontinuity at $\pm\pi$~\cite{zhou2019continuity}, it barely poses a problem since rotations between two views are often far from $\pm\pi$ (such strong rotation will quickly diminish the overlapping field of view). We observe similar performance from the two representations, but opt for Euler angels since its fusion of yaw-pitch-roll angles, denoted as $\boldsymbol{\theta}_R=\{\phi_y,\phi_p,\phi_r\}$, has a clearer geometric meaning.

\noindent{\textbf{Loss}~~~} One could impose the loss directly on the fused angles $\{\boldsymbol{\theta}_{R,f},\boldsymbol{\theta}_{t,f}\}$ with a circular loss~\cite{circularloss}, or convert them back to $(\boldsymbol{R}(\boldsymbol{\theta}_{R,f}),\boldsymbol{t}(\boldsymbol{\theta}_{t,f}))$ and impose loss on it. All the combinations perform similarly in our experiments; we choose the following loss for marginally better performance,
\begin{equation}
L(\boldsymbol{\theta}_{R,f},\boldsymbol{\theta}_{t,f}) = |\boldsymbol{t}(\boldsymbol{\theta}_{t,f})-\boldsymbol{t}^*|_1 + w|\boldsymbol{\theta}_{R,f}-\bar{\boldsymbol{\theta}}_R^*|_1,
\end{equation}
where $*$ indicates the ground truth and $\bar{\boldsymbol{\theta}}_R^*$ is ${\boldsymbol{\theta}}_R^*$'s circular nearest neighbour to $\boldsymbol{\theta}_{R,f}$. $w=1.0$ is a weight.

\noindent{\textbf{Implementation details}~~~}
We make use of OpenCV for the 5-point solver with RANSAC, and Ceres~\cite{agarwal2012ceres} for BA and Jacobian computation. %
More discussions on the framework and network training are in the supplementary.

\section{Experiments}

 \begin{table*}\small
 		\begin{center}
 		\begin{tabular}[H]{|c|c|c||c|c||c|c||c|c|}
 			\hline
 			 & \multicolumn{2}{c||}{{MVS}} & \multicolumn{2}{c||}{{Scenes11}} & \multicolumn{2}{c||}{{RGB-D}} & \multicolumn{2}{c|}{{Sun3D}}\\
 			\cline{2-9}
 			  &  Rot. & Tran. & Rot.& Tran.& Rot. & Tran.& Rot.& Tran.\\
 			\hline
 			DeMoN~\cite{ummenhofer2017demon}  &  5.156 & 14.447 & 0.809 & 8.918 & 2.641 & 20.585& 1.801 & 18.811 \\
 			\hline
 			LS-Net~\cite{clark2018ls} &  4.653& 11.221 &4.653 & 8.210 & \textbf{1.010} & 22.110 & 1.521 & 14.347\\
 			\hline
 			BA-Net~\cite{tang2018ba} &  3.499& 11.238 & 3.499 & 10.370 & 2.459 & 14.900 & 1.729 & 13.260\\
 			\hline
 			DeepSfM~\cite{wei2019deepsfm} & 2.824& 9.881 &0.403 & 5.828 & 1.862 & 14.570 & 1.704 & 13.107\\
 			\hline
 			LGC-Net~\cite{yi2018learning} &  2.753& 3.548 & 0.977 & 4.861 & 2.014 & 16.426 & 1.386 & 14.118 \\
 			\hline
 			NM-Net~\cite{zhao2019nm} & 6.628& 12.595 &15.717 & 31.477 & 13.444 & 34.212 & 4.393 & 21.091 \\
 			\hline
 			SIFT+5pt\&BA & 1.313 & 2.555 & 2.062 & 8.125 & 6.895 & 29.457 & 2.516 & 21.925  \\
 			\hline
 			UA-Fusion-SIFT & \textbf{1.203}  & \textbf{2.403} & 0.525 & 4.322 & 2.274 & 15.570  & 1.960 & 17.340\\
 			\hline
 			SuperGlue+5pt\&BA & 2.884 & 5.024 & 0.390 & 3.872 & 1.829 & 16.330 & \textbf{1.255} & 12.200 \\
 			\hline
 			UA-Fusion & 2.502  & 4.506 & \textbf{0.388} & \textbf{3.001} & 1.480 & \textbf{10.520}  & 1.340 & \textbf{11.830}\\
 			\hline
 		\end{tabular} 
 		\end{center}
 		\caption{\textbf{Quantitative comparison with state-of-the-art methods on DeMoN datasets.} Both translation (Tran.) and Rotation (Rot) errors are measured in degree ($\circ$). The lowest error in each column is bolded.}
 	 	\label{tab:demoncomparison}	
 \end{table*}

\begin{figure*}
 \vspace{-0.1cm}
\centering
\begin{minipage}[c]{0.30\textwidth}
\centering
\includegraphics[width=0.85\linewidth, trim = 0mm 5mm 0mm 20mm, clip]{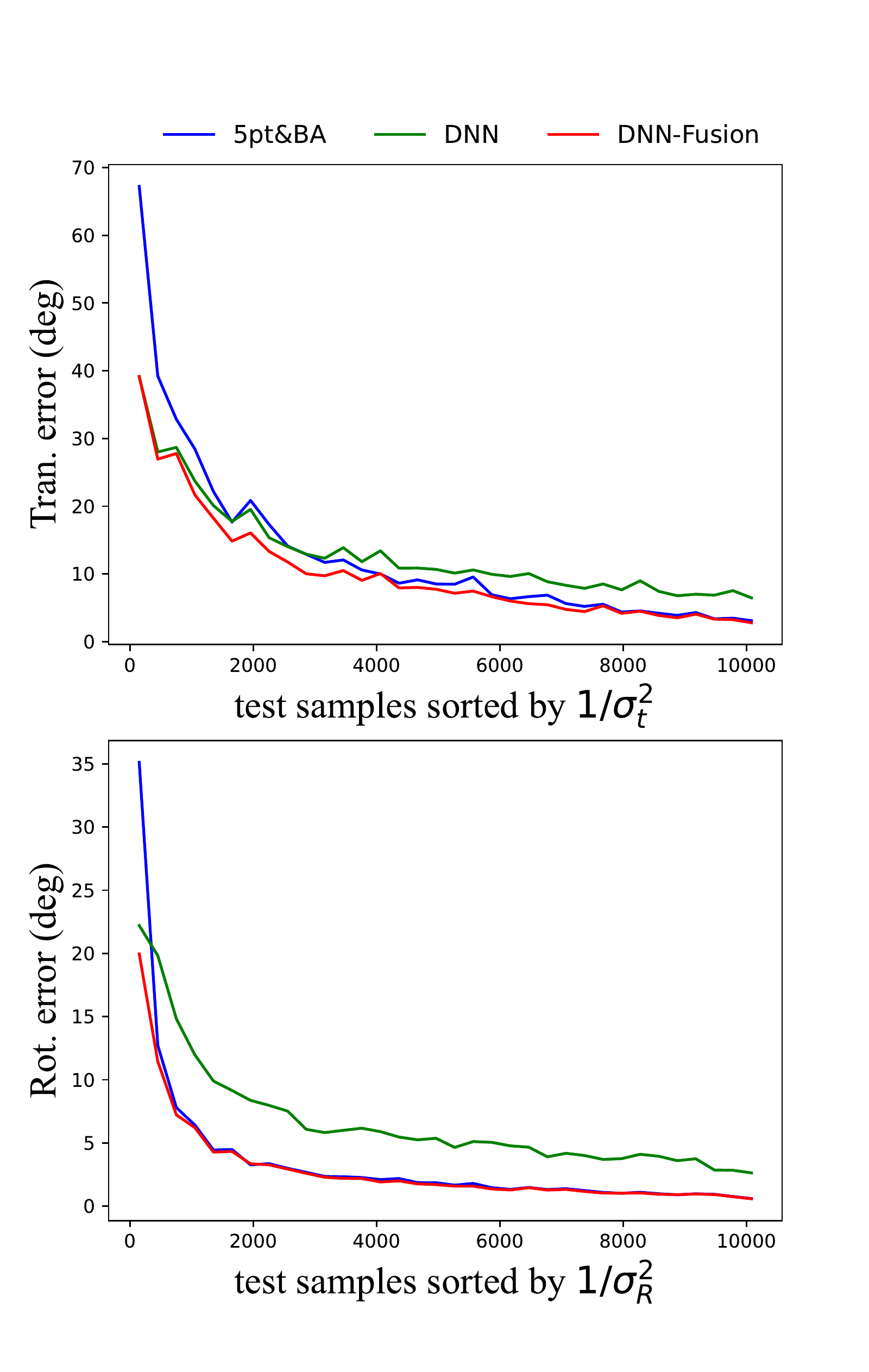}
\caption{\textbf{Error comparison} between geometric and DNN predictions.}
\label{fig:errorcurve}
\end{minipage}
\hspace{0.7cm}
\begin{minipage}[c]{0.65\textwidth}
\centering
\includegraphics[width=1.0\linewidth, trim = 40mm 10mm 45mm 10mm, clip]{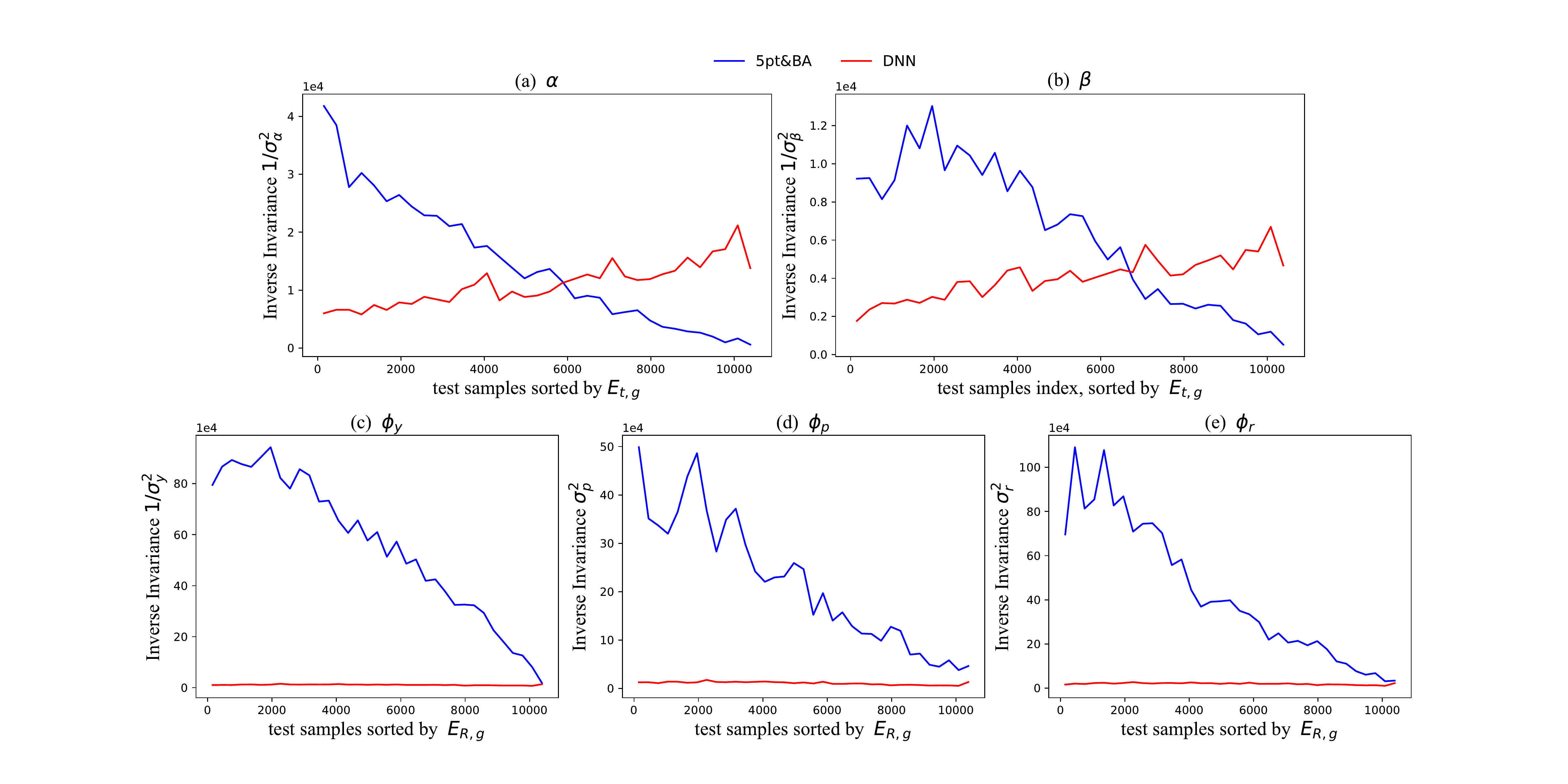}
\caption{Comparison between the \textbf{geometric and DNN uncertainty} for each parameter in $\{\boldsymbol{\theta_t},\boldsymbol{\theta_R}\}$. }
\label{fig:uncertaintycurve}
\end{minipage}
\end{figure*}

\subsection{Dataset}
\noindent \textbf{DeMoN Dataset}~\cite{ummenhofer2017demon}. This dataset contains a large amount of data including indoor, outdoor, and synthetic data, extracted from SUN3D~\cite{xiao2013sun3d}, RGB-D~\cite{sturm12iros}, MVS~\cite{ummenhofer2015global,fuhrmann2014mve,schonberger2016pixelwise,schonberger2016structure}, and ShapeNet~\cite{chang2015shapenet}. The diversity of both scenes and motions makes the dataset challenging for SFM.

\noindent \textbf{ScanNet}~\cite{dai2017scannet}. To evaluate the generalization ability, we also test our network on the ScanNet data. We use the testing data extracted by~\cite{tang2018ba}, that consists of 2000 pairs of images with accurate ground truth pose. The sequences are captured indoors by hand-held cameras, making it particularly hard.

\subsection{Results on DeMoN}
We first compare UA-Fusion with the state-of-the-art relative pose learning methods including DeMoN~\cite{ummenhofer2017demon}, LS-Net~\cite{clark2018ls}, BA-Net~\cite{tang2018ba} and DeepSfM~\cite{wei2019deepsfm}, in Tab.~\ref{tab:demoncomparison}. 
The error of translation (resp. rotation) is measured as the angle between the prediction and the ground truth.  
First, although SuperGlue followed by 5pt\&BA produces competitive results compared to methods that learn pose directly,
it can be seen that our fusion of the geometric and DNN prediction further boosts the performance, achieving the overall best results. We also test UA-Fusion with SIFT as the feature matcher, under which case ours again improves over the pure geometric method, further validating its merits. In addition, we test two more methods that focus on learning correspondences, LGC-Net~\cite{yi2018learning} and NM-Net~\cite{zhao2019nm}, by passing their output to 5pt\&BA. We observe that  the obtained accuracies lag behind our fusion method.

\subsection{Analysis: Geometric vs. DNN}
\label{sec:analysis}

\newcommand{\tabincell}[2]{\begin{tabular}{@{}#1@{}}#2\end{tabular}}
\begin{figure*}
\vspace{-0.1cm}
  \begin{minipage}[c]{0.60\linewidth}
    \centering 
    \includegraphics[width=0.95\linewidth, trim = 45mm 10mm 40mm 10mm, clip]{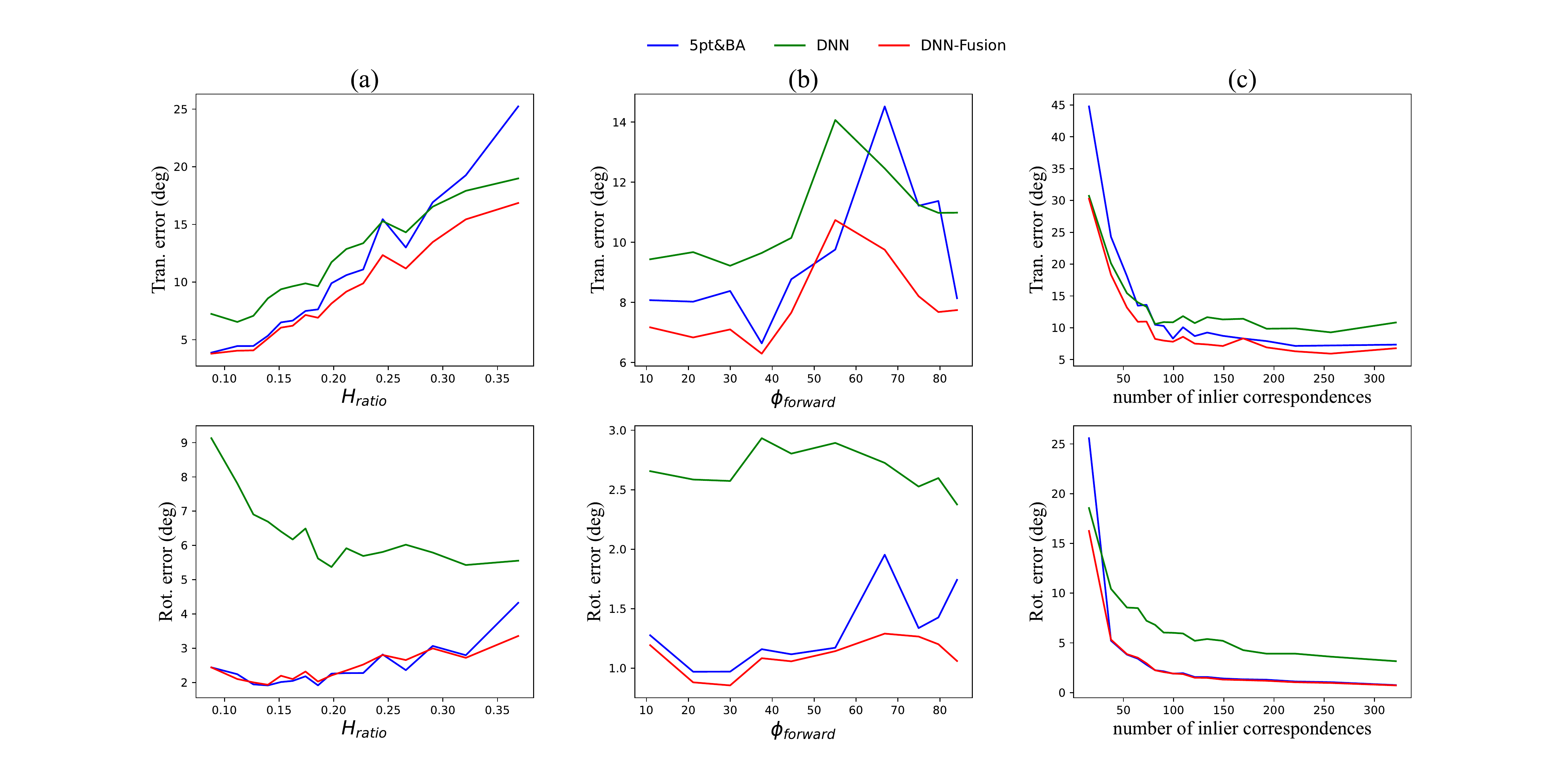} 
    \caption{Comparisons of errors against (a) \textbf{homography degeneracy}, (b) \textbf{forward and sideway motion}, and (c) \textbf{correspondences}. } 
    \label{fig:error_homo_forward_inlier} 
  \end{minipage}%
  \hspace{0.7cm}
  \begin{minipage}[c]{0.35\linewidth} 
    \begin{center}
    \begin{tabular}[H]{|@{ }c@{ }|c|c|}
 			\hline
 			ScanNet & Rot. & Tran. \\
 			\hline
 			DeMoN~\cite{ummenhofer2017demon} & 3.791 & 31.626  \\
 			\hline
 			LSD-SLAM~\cite{engel2014lsd} & 4.409  & 34.360 \\
 			\hline
 			BA-Net~\cite{tang2018ba} & 1.587 & 31.005 \\
 			\hline
 			DeepSfM~\cite{wei2019deepsfm}  & 1.588  & 30.613 \\
 			\hline
 			\tabincell{c}{SuperGlue+5pt\&BA\\ (indoor)}   & 1.776 & 20.936 \\
 			\hline
 			UA-Fusion (indoor)  & \textbf{0.814} & \textbf{16.517} \\
 			\hline
 			\tabincell{c}{SuperGlue+5pt\&BA\\ (outdoor)} & 2.519 & 20.470  \\
 			\hline
 			UA-Fusion (outdoor) & 0.824 & 17.495 \\
 			\hline
 	\end{tabular} 
 	\end{center}
    \makeatletter\def\@captype{table}\makeatother\caption{\textbf{Quantitative comparisons on ScanNet}. Both indoor and outdoor SuperGlue model are tested.}
    \label{tab:scannet} 
  \end{minipage} 
\end{figure*}

Since the DeMoN test set contains only 354 pairs of images, we train a model with a few training sequences ($\sim$10k pairs) left for testing, in order for statistically meaningful analysis.
We denote the translation and rotation error of the geometric solution (``5pt\&BA"), the plain DNN predition prior to fusion (``DNN"), and the final fused solution (``DNN-Fusion") as $(E_{t,g},E_{R,g})$, $(E_{t,d},E_{R,d})$ and $(E_{t,f},E_{R,f})$.

\noindent \textbf{Geometric error vs. DNN error~~~}We first study the accuracy against the uncertainty of the geometrical solution. We sort the test data by the total inverse variance in translation $1/\sigma_{t}^{2}=1/\sigma_{\alpha}^{2}+1/\sigma_{\beta}^{2}$, and plot the sorted $E_{t,g}$, $E_{t,d}$ and $E_{t,f}$ in the top row of Fig.~\ref{fig:errorcurve}. The curves are smoothed by averaging over every 300 consecutive image pairs. First observe the overall trend that $E_{t,g}$ decreases while $1/\sigma_{t}^{2}$ increases, revealing the correlation between geoemtrical error and uncertainty. In addition, the accuracy gap between $E_{t,g}$ and $E_{t,f}$ is increasingly large with decreasing $1/\sigma_{t}^{2}$. This indicates a more significant role of DNNs in geometrically uncertain scenarios; conversely, the geometrical solution is mostly retained if it is highly confident. Further observe the superiority of ``DNN-Fusion" over ``DNN", which is expected as only ``DNN-Fusion" receives direct supervision. Another potential reason is that ``DNN" likely has to contain bias to compensate the error/bias in ``5pt\&BA" during fusion. The same curve is plotted for the rotation error in the bottom row of Fig.~\ref{fig:errorcurve}. We observe that the geometrical solution transfers to the final output in most cases, probably due to the higher stability of the rotation estimation compared to the translation; this viewpoint has been conveyed  by Nist$\acute{\text{e}}$r~\cite{nister2004efficient}.

\noindent  \textbf{Geometric uncertainty vs. DNN uncertainty~~~}Next, we visualize the uncertainty by plotting $(1/\sigma_\alpha^{2},1/\sigma_\beta^{2})$ sorted by $E_{t,g}$. As shown in Fig.~\ref{fig:uncertaintycurve}(a)(b), $(1/\sigma_{\alpha,d}^{2},1/\sigma_{\beta,d}^{2})$ are higher than $(1/\sigma_{\alpha,g}^{2},1/\sigma_{\beta,g}^{2})$ for cases with larger $E_{t,g}$, and vice verse. This implies the desired capability of ``DNN" to dominate the final solution when ``5pt\&BA" tends to fail. Similarly, we present the same plot for rotation in Fig.~\ref{fig:uncertaintycurve}(c)-(e). 
Comparing the range of $1/\sigma_{R,g}^{2}$ and $1/\sigma_{t,g}^{2}$, one observes the higher confidence/stabability in the geometric rotation estimate than the translation. This also makes it dominate in most cases when fused with the DNN prediction, as evidenced by the curves. 
We also note that the smoothed curves reveal the overall trend but may overly obscure the DNN's impact on the rotation, 
thus we provide in the supplementary more analyses in this regard.

\noindent \textbf{Homography degeneracy~~~}
Motion estimation may be unstable with correspondences that could be related by a homography.
Characterizations of the algorithm's stability in this aspect is essential. To this end, we fit a homography to correspondences in each image pair; the ratio of inliers, $H_{ratio}$, is used to indicate closeness to degeneracy. We then sort image pairs by $H_{ratio}$ and plot the error curves in Fig.~\ref{fig:error_homo_forward_inlier}(a). As can be seen, ``DNN-Fusion" mostly keeps the geometric solution in well-conditioned cases with lower $H_{ratio}$, while alleviates the degradation under higher $H_{ratio}$.

\noindent  \textbf{Forward vs. Sideway motion~~~}
We first select those image pairs with nearly pure translational motion, and compute the angle between translation direction and z-axis, denoted as $\phi_{forward}$. $\phi_{forward}=0^\circ$ and $90^\circ$ indicate pure forward and sideway motion, respectively. We plot the errors sorted by $\phi_{forward}$ in Fig.~\ref{fig:error_homo_forward_inlier}(b). While forward motion does not cause serious issues in our tested data, one observes increasing errors while approaching pure sideway motion, which is liable to the bas-relif ambiguity. Yet, the performance degradation is alleviated in our fused solution.

\noindent  \textbf{Correspondence~~~}
Finally, we plot in Fig.~\ref{fig:error_homo_forward_inlier}(c) the errors against the number of inlier correspondences found by ``5pt\&BA". Clearly, DNNs contribute more when the geometric accuracy drops due to lack of correspondences.

\begin{figure}
	\centering
	    \vspace{-0.1cm}
		\includegraphics[width=0.85\linewidth, trim = 0mm 0mm 190mm 0mm, clip]{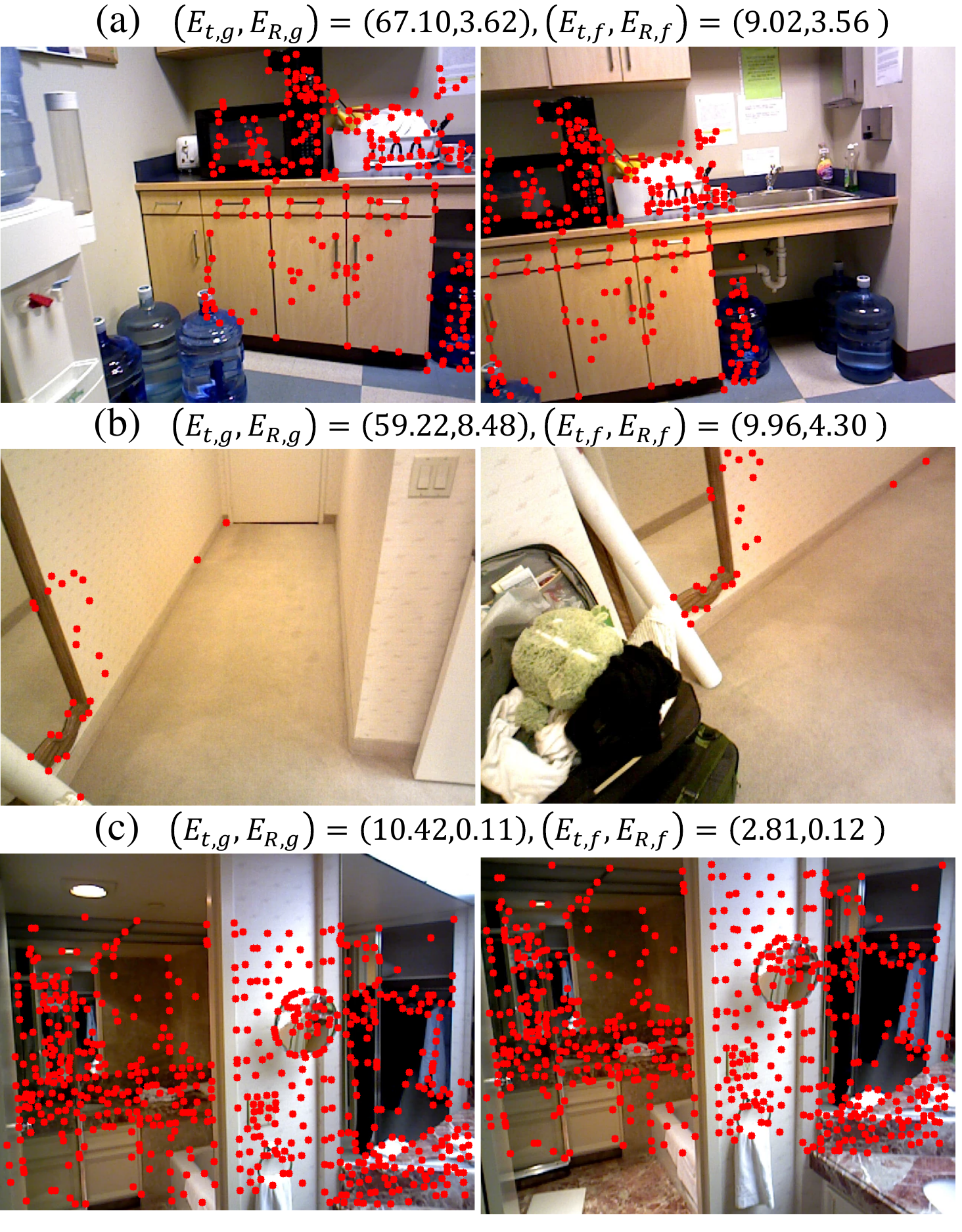}
	\centering
	\caption{\textbf{Qualitative results} with the geometric and DNN errors on top of each image pair. Red dots mark the correspondences.}
	\label{fig:qualitativeexample}
\end{figure}

\noindent \textbf{Qualitative Results~~~}We present here three qualitative examples. Fig.~\ref{fig:qualitativeexample}(a) represents a geometrically challenging case since all the correspondences nearly lie in a frontal plane with small depth variations, leading to weak perspective effect and potential degeneracy.  Similarly, Fig.~\ref{fig:qualitativeexample}(b) shows an ill-conditioned instance with sparse correspondences concentrated on a small region. Fig.~\ref{fig:qualitativeexample}(c) demonstrates a case with nearly lateral motion along the vertical direction, which is liable to bas-relief ambiguity. As can be seen, the geometric method may  deteriorate significantly when confronted with such challenges, especially the translation, whereas our fusion network returns more reasonable solutions.

\begin{figure*} 
 \vspace{-0.cm}
  \begin{minipage}[c]{0.26\linewidth}
    \centering 
    \includegraphics[width=1.0\linewidth, trim = 20mm 0mm 160mm 14mm, clip]{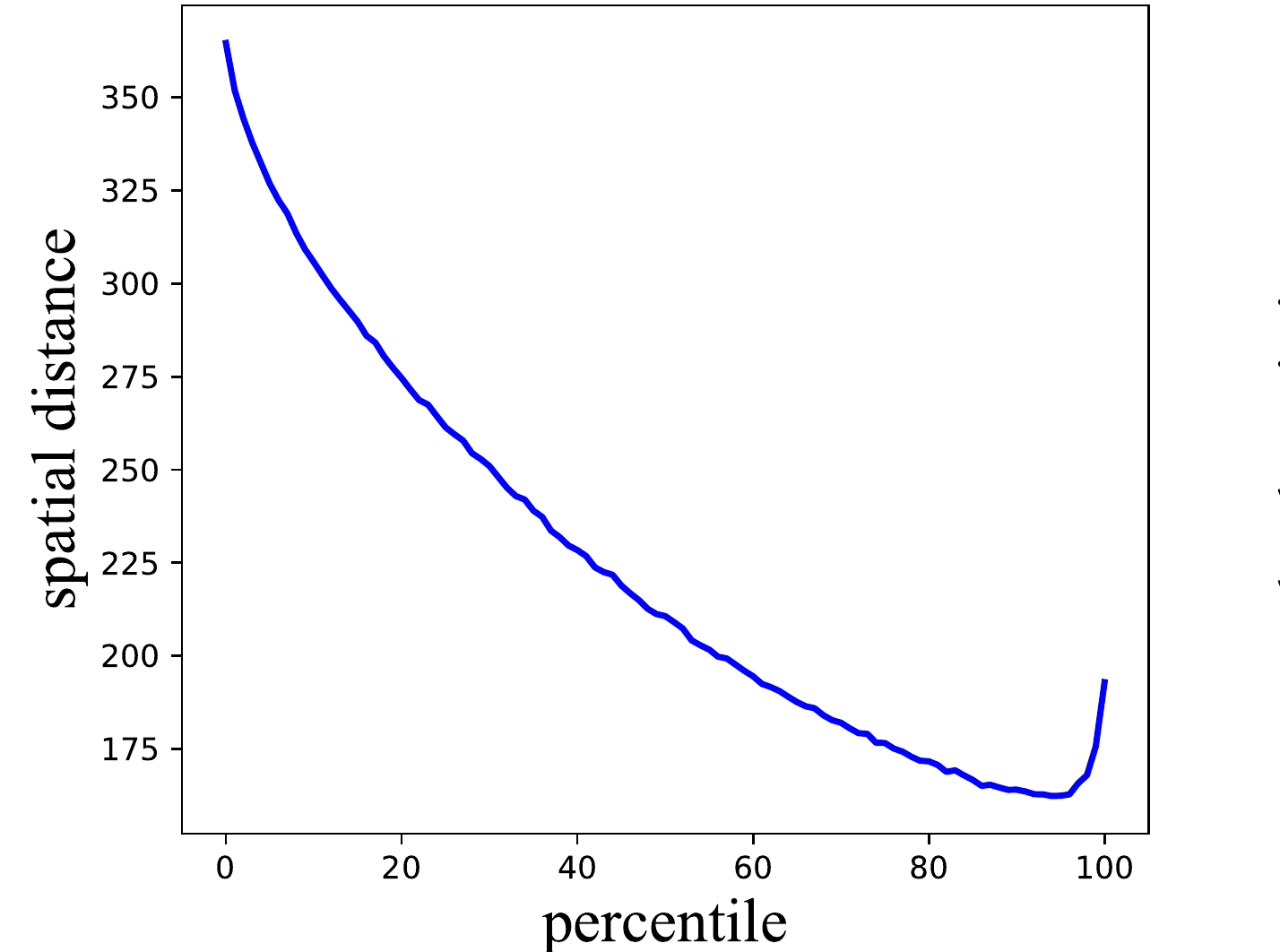} 
    \caption{Study of spatial distances (pixels) against \textbf{attentions}.} 
    \label{fig:attention.eps} 
  \end{minipage}%
  \hspace{0.1cm}
  \begin{minipage}[c]{0.73\linewidth} 
  \setlength{\tabcolsep}{4pt}
    \begin{center}
    \scalebox{0.85}{
 		\begin{tabular}[t]{|@{ }c@{ }|c|c|c|c|c|c|c|c|c|c|}
 			\hline
 			 & \multicolumn{2}{c|}{{MVS}} & \multicolumn{2}{c|}{{Scenes11}} & \multicolumn{2}{c|}{{RGB-D}} & \multicolumn{2}{c|}{{Sun3D}} & \multicolumn{2}{c|}{{All}}\\
 			\cline{2-11}
 			  &  Rot. & Tran. & Rot.& Tran.& Rot. & Tran.& Rot.& Tran. & Rot.& Tran.\\
 			\hline
 			Aleatoric Uncertainty & 2.890  & 4.954 & \textbf{0.375} & 2.956 & 1.813 & 13.650  & 1.273 & 11.750 & 1.371 & 7.694 \\
 			
 			\hline
 			w/o Uncertainty  & 4.608  & 7.233 & 1.034 & 5.924  & 4.121 & 12.400 & 2.479 & 13.870 & 2.712 &9.200 \\
 			\hline
 			Median Uncertainty & 3.638 & 4.841 & 0.764 & 4.262 & 2.080 & 10.960 & 1.669& 13.200 & 1.790 & 7.838\\
 			\hline
 			w/o Attention &  2.956 & 6.051 &  0.410 & 3.237   &1.678 & 12.65  & 1.321 & 12.14 & 1.380 & 7.870\\
 			\hline
 			Discontinuity  &  2.910 & \textbf{3.984} & 0.384 & 3.268 & 1.667& 12.160 & 2.298 & 12.310 & 1.573 & 7.452\\
 			\hline
 			w/o ResNet    & 2.519 & 4.539  & 0.409 & \textbf{2.805}   & 1.742& 13.550  & 1.261 & \textbf{11.630} & 1.306& 7.506\\
 			\hline
 			w/o GNN & 3.035 & 6.080 & 0.399 & 3.266  &1.826 & 14.220   & \textbf{1.255} & 13.550 & 1.406 & 8.589\\
 			\hline
 			PointNet  & 2.833 & 5.665  & 0.409 & {2.850}  & 1.652 & 13.220  & 1.270 & 13.490 & 1.333 &8.080 \\
 			\hline
 			UA-Fusion & \textbf{2.502}  & 4.506 & 0.388 & 3.001 & \textbf{1.480} & \textbf{10.520}  & 1.340 & 11.830 &\textbf{1.246} & \textbf{6.888}\\
 			\hline
 		\end{tabular} }
 		\end{center}
 		\makeatletter\def\@captype{table}\makeatother\caption{\textbf{Ablation study} of our methods under different configurations. We also compute the averaged error over all the four scenes, as shown in the ``All" column. The lowest error in each column is bolded.}
 	 	\label{tab:ablation}
  \end{minipage} 
\end{figure*}

\subsection{Self-attention}
A comprehensive understanding on what is learned by the attention module is challenging. However, as discussed in Sec.~\ref{sec:architecture}, we could study its potential relation to the spatial distance between any two pairs of correspondences. To this end, for a pair of correspondences, we collect its attentions to all the other pairs of correspondences in the last self-attention layer, and then compute its spatial distance\footnote{Denoting a pair of correspondences as $(\boldsymbol{x}_i^1,\boldsymbol{x}_i^2)$ and $(\boldsymbol{x}_j^1,\boldsymbol{x}_j^2)$, the spatial distance is defined as$(\|(\boldsymbol{x}_i^1-\boldsymbol{x}_j^1)\|+\|(\boldsymbol{x}_i^2-\boldsymbol{x}_j^2)\|)/2$.} to other correspondences at different percentiles of the attention set. This way we can compare spatial distance against varying attention. Averaging over all the points in the entire testing set, we obtained the curve shown in Fig.~\ref{fig:attention.eps}. First, in line with the expectation, the overall trend indicates strong correlation between attention and spatial distance. In addition, it also reveals the overall smaller spatial distance with increasingly higher attentions. Referring to Sec.~\ref{sec:architecture}, we reckon that this is due to the increasing difficulty of extracting additional pose-related information from two points closer to each other; such difficulty enforces stronger interactions between those points in order to make contributions to the final camera pose estimation. More discussions are in the supplementary due to lack of space.

\subsection{Baselines \& Ablation Study}
We provide ablation studies to elucidate the impact of important factors by comparison with several baselines.

\noindent \textbf{Aleatoric Uncertainty}: We replace our geometry-guided uncertainty with the plain aleatoric uncertainty~\cite{kendall2017uncertainties,klodt2018supervising,laidlow2019deepfusion}, and leverage fusion as a post-processing step.

\noindent \textbf{w/o Uncertainty}: We remove the uncertainty head and the fusion step, only regressing the camera pose. 

\noindent \textbf{Median Uncertainty}: Instead of predicting uncertainty by the network, we simply assign each $\sigma_d^{-1}$ a constant value---the median of $\sigma_g^{-1}$'s in the entire training set. This way, the DNN prediction would dominate the final solution on half of the data with highest geometric uncertainty. 

\noindent \textbf{w/o Attention}: We remove the self-attention mechanism and use plain MLPs to compute the messages. 

\noindent \textbf{Discontinuity}: In this case, the network directly predicts $(\alpha,\beta)$ instead of $(t_x,t_y,t_z)$, as discussed in Sec.~\ref{sec:motionparameter}.

\noindent \textbf{w/o ResNet} and \textbf{w/o GNN}: Here, we remove either ResNet or self-attention GNN from the network, i.e. relying on the geometric feature or appearance feature alone. 

\noindent \textbf{PointNet}: We replace the self-attention GNN with PointNet~\cite{qi2017pointnet}, which has a similar number of parameters.

As shown in Tab.~\ref{tab:ablation}, all the baseline methods lead to a drop in the accuracy. This substantiates the effectiveness of our geometry-guided uncertainty, self-attention mechanism, motion prameterization, and network design. In particular, removing uncertainty and fusion causes the most significant degradation. We present more analysis on the aleatoric uncertainty in the supplementary due to lack of space.


\subsection{Results on ScanNet}
To demonstrate the generalization capability of our network trained on the DeMoN datasets, we also conduct cross-validation experiments on the ScanNet dataset. Since the indoor model of SuperGlue is trained on ScanNet as well and its training set may have intersection with the testing set here, we instead apply its outdoor model to obtain correspondences. As a reference, we also report the results with the indoor model. As can been in Tab.~\ref{tab:scannet}, UA-Fusion achieves the best accuracy compared to prior arts. More importantly, we observe similar behavior in accuracy and uncertainty as analyzed in Sec.~\ref{sec:analysis}; this is detailed in the supplementary.

\vspace{0.1cm}
\noindent{\textbf{Computational performance:}} Our network inference takes about 0.01s per image pair, on an RTX2080 GPU. Computing geometric uncertainty takes around 0.025s, where the block sparsity of the Jacobian is leveraged for efficiency.

\section{Conclusion}

In this paper, we propose a new relative pose estimation framework capable of reaping benefits from both the classical geometric methods and deep learning approaches. At the crux of our method is learning the uncertainty explicitly governed by the geometric solution, which permits fusion in a probabilistic manner during training. Despite being specific to the relative pose problem, we envision that our idea of learning with geometric uncertainty guidance followed by fusion in a deep network is broadly applicable.

{\small
\bibliographystyle{ieee_fullname}
\bibliography{egbib}
}

\end{document}